%% file: main.tex
\documentclass[sigconf, natbib, nonacm]{acmart}

\AtBeginDocument{%
  }

\acmISBN{}

\usepackage{array}
\usepackage{subcaption}
\usepackage{tabularx}
\usepackage{enumitem}
\setlist[enumerate]{leftmargin=*}
\newcolumntype{L}[1]{>{\raggedright\arraybackslash}m{#1}}

\begin{document}

\title{When More Is Less: A Systematic Analysis of Spatial and Commonsense Information for Visual Spatial Reasoning}
\author{Muku Akasaka}
\email{akasakam@student.unimelb.edu.au}
\affiliation{%
  \institution{The University of Melbourne}
  \city{Melbourne}
  \state{Victoria}
  \country{Australia}
}

\author{Soyeon Caren Han}
\email{caren.han@unimelb.edu.au}
\affiliation{%
  \institution{The University of Melbourne}
  \city{Melbourne}
  \state{Victoria}
  \country{Australia}
}

\begin{abstract}
Visual spatial reasoning (VSR) remains challenging for modern vision-language models (VLMs), despite advances in multimodal architectures. A common strategy is to inject additional information at inference time, such as explicit spatial cues, external commonsense knowledge, or chain-of-thought (CoT) reasoning instructions. However, it remains unclear when such information genuinely improves reasoning and when it introduces noise. 
In this paper, we conduct a hypothesis-driven analysis of information injection for VSR across three representative VLMs and two public benchmarks. We examine (i) the type and number of spatial contexts, (ii) the amount and relevance of injected commonsense knowledge, and (iii) the interaction between spatial grounding and CoT prompting. Our results reveal a consistent pattern: more information does not necessarily yield better reasoning. Targeted single spatial cues outperform multi-context aggregation, excessive or weakly relevant commonsense knowledge degrades performance, and CoT prompting improves accuracy only when spatial grounding is sufficiently precise.
These findings highlight the importance of selective, task-aligned information injection and provide practical guidance for designing reliable multimodal reasoning pipelines.
\end{abstract}

\maketitle

\begingroup
\renewcommand{\thefootnote}{}
\footnotetext{\textcopyright\ 2026 Muku Akasaka and Soyeon Caren Han.
This is the authors' submitted version of a work submitted to the 49th International ACM SIGIR Conference on Research and Development in Information Retrieval (SIGIR '26).
It is posted here for personal use. Not for redistribution.
If the work is published in the ACM Digital Library, the Version of Record link (including the DOI) will be added to this record.}
\addtocounter{footnote}{-1}
\endgroup

\input{sections/01_introduction}

\input{sections/02_hypotheses}

\input{sections/03_intervention}
\input{sections/04_setup}

\input{sections/05_results}

\input{sections/06_conclusion}


\bibliographystyle{ACM-Reference-Format}
\bibliography{bibliography}

\appendix



\end{document}

%% file: sections/01_introduction.tex
\section{Introduction}
Visual spatial reasoning (VSR), understanding relations such as left/right, front/back, overlap, and proximity, remains a persistent challenge for vision-language models (VLMs)~\cite{zellers2019recognition, liu2023visual}. Although recent multimodal systems demonstrate strong performance on general image-text tasks, they often struggle when spatial relations are ambiguous, frame-dependent, or require precise grounding.
Recent efforts address this limitation either through spatially-aware architectures and specialised datasets \cite{chen2024spatialvlm, cheng2024spatialrgpt}, or by injecting external spatial cues at inference time without additional training \cite{wang2024orient, jia2025omnispatial}. In this work, we focus on the inference-time setting and augment inputs with explicit spatial contexts, relational commonsense knowledge, or chain-of-thought (CoT) reasoning instructions~\cite{wei2022chain}.
Intuitively, providing richer input should facilitate better reasoning.
However, additional information does not consistently improve performance.
Injected context may be redundant, weakly aligned with the task, or difficult to integrate, and structured reasoning prompts can either clarify inference or amplify incorrect assumptions.
These observations raise a central question:
\emph{When does information injection genuinely improve visual spatial reasoning, and when does it become harmful?}

We address this question through a hypothesis-driven empirical analysis.
Rather than proposing a new architecture, we treat VLMs as fixed black-box systems and systematically vary the type and quantity of injected information.
We evaluate three representative VLMs (Qwen-2-VL~\cite{Qwen2VL}, LLaVA-Next-1.6~\cite{liu2024llavanext}, and BLIP-3~\cite{xue2024xgenmmblip3familyopen}) on two public VSR benchmarks~\cite{liu2023visual, du-etal-2024-embspatial} and test three hypotheses: (1) Not all spatial contexts contribute equally; single, task-relevant cues outperform multi-context aggregation. (2) Excessive or weakly relevant commonsense knowledge degrades reasoning due to information overload.
(3) CoT prompting improves performance only when spatial grounding is sufficiently precise.
Importantly, improvements obtained through information injection can mask fragile reasoning behaviour.
A model may appear more capable simply because it leverages injected cues heuristically, rather than developing robust spatial grounding.
Without understanding how different forms of information interact with internal representations, it is difficult to assess whether performance gains reflect genuine improvements in reasoning or merely the exploitation of superficial shortcuts.
Across models and datasets, we observe a consistent pattern: more information does not necessarily lead to better reasoning.
Effective VSR requires selective, task-aligned grounding rather than indiscriminate context accumulation.
Our findings provide practical guidance for designing and evaluating multimodal reasoning pipelines and highlight the importance of representation and information control in VLM-based reasoning. Main contributions are as follows:
\begin{enumerate}
    \item A hypothesis-driven empirical analysis of information injection for visual spatial reasoning across three representative VLM families and two public benchmarks.
    \item Systematic evaluation of spatial context type, context aggregation, commonsense relevance threshold, and CoT prompting under controlled input interventions.
    \item Empirical evidence that more information does not necessarily improve reasoning, revealing diminishing returns, overload effects, and grounding-dependent CoT behaviour.
    \item Practical insights for designing reliable multimodal prompting strategies through selective, task-aligned information injection.
\end{enumerate}

%% file: sections/02_hypotheses.tex
\section{Hypotheses}
\label{sec:hypo}
Recent VLMs still struggle with VSR, and a common practice is to inject additional information, including spatial cues, commonsense knowledge, or chain-of-thought (CoT) instructions, to compensate for missing grounding or reasoning capabilities. However, more information does not necessarily translate to better reasoning: added context can be ignored, misused, or even act as noise. In this paper, we formulate three hypotheses to systematically characterise \emph{when} injected information helps and \emph{when} it harms.

\textbf{H1: Not all spatial contexts contribute equally to VSR.}
Single, targeted spatial cues yield larger and more reliable gains than aggregating multiple spatial contexts.
We expect (i) substantial variance across different spatial context types, and (ii) diminishing returns or degradation as more contexts are appended, due to increased integration burden.

\textbf{H2: Excessive or weakly relevant commonsense knowledge can degrade VSR.}
Commonsense injection is beneficial only when it is both relevant and concise; otherwise, it introduces distraction and overwhelms reasoning.
We operationalise this hypothesis by varying both (i) the relevance threshold controlling the amount/quality of retrieved knowledge, and (ii) the knowledge type, and test whether performance peaks at selective settings rather than at maximal injection.

\textbf{H3: Chain-of-Thought helps only when spatial grounding is sufficiently precise.}
CoT-style prompting improves VSR when the model has reliable premises (i.e., accurate grounding and unambiguous inference), but can amplify errors when grounding is uncertain, or the task is frame-dependent.
We test this primarily through qualitative comparisons, complemented by quantitative trends observed across settings.

%% file: sections/03_intervention.tex
\begin{figure}[t]
    \centering    \includegraphics[width=\columnwidth]{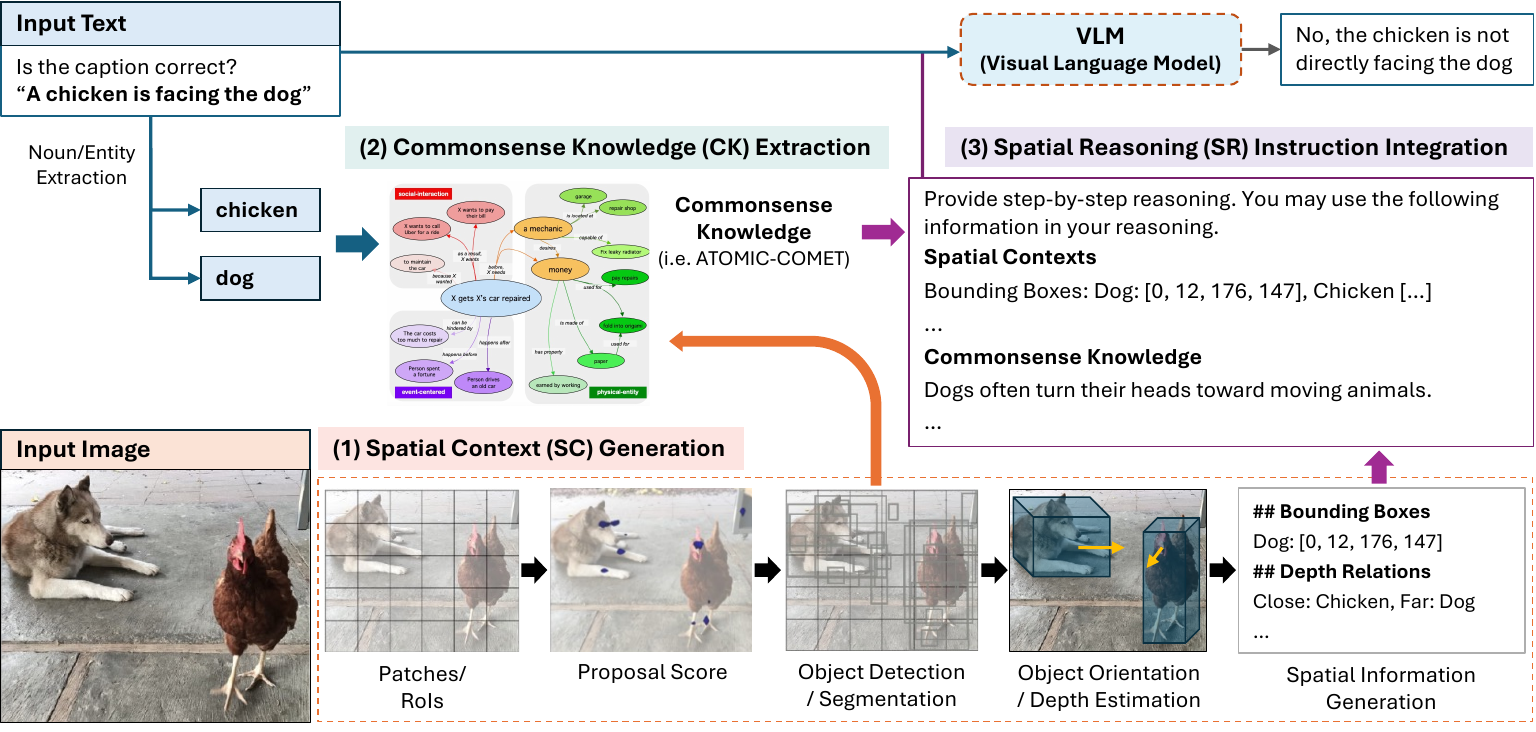}
    \caption{Overview of the controlled input intervention setup. We treat VLMs as black-box systems and vary only the injected information: (1) Spatial Contexts (SC), (2) Commonsense Knowledge (CK) retrieved from a knowledge base, and (3) Spatial Reasoning instructions (SR).}
    \label{fig:overview}
    \vspace{-2em}
\end{figure}

\section{Intervention Setup: Injected Information}
\label{sec:intervention}
We analyse how external information affects visual spatial reasoning in VLMs by treating each model as a black-box image--text interface and varying only the \emph{input} conditions. Given an image and a question, we optionally append (i) explicit spatial contexts extracted by off-the-shelf vision modules, (ii) relational commonsense statements retrieved from a knowledge base, and (iii) chain-of-thought style reasoning instructions. The following types of information are used in our interventions.

\noindent\textbf{Spatial Contexts (SC).}
We extract three spatial values and six interpretable spatial contexts that describe pairwise relations or object attributes: bounding boxes, orientation angles, metric depth values, lateral (left/right), vertical (above/below), orientation (facing direction), depth (close/far), overlap (occlusion-like), and size (large/small). These contexts are provided in a structured format.

\noindent\textbf{Commonsense Knowledge (CK).}
To study relational priors beyond geometry, we retrieve short commonsense statements from ATOMIC$^{20}_{20}$ that describe plausible interactions between detected entities.
Candidate statements are selected based on their semantic similarity to the caption and detected entities.
We control the amount and relevance of injected knowledge using a similarity threshold and knowledge type (PE/EC/SI).

\noindent\textbf{Spatial Reasoning Instructions (SR).}
We prepend brief step-by-step reasoning instructions that encourage the model to use the provided spatial and commonsense information before answering.

Across all experiments, LVLM parameters remain fixed; only these input conditions are varied to isolate the causal effect of each information type.

%% file: sections/04_setup.tex
\section{Evaluation Setup}
\begin{figure}[t]
  \centering
  \includegraphics[width=\linewidth]{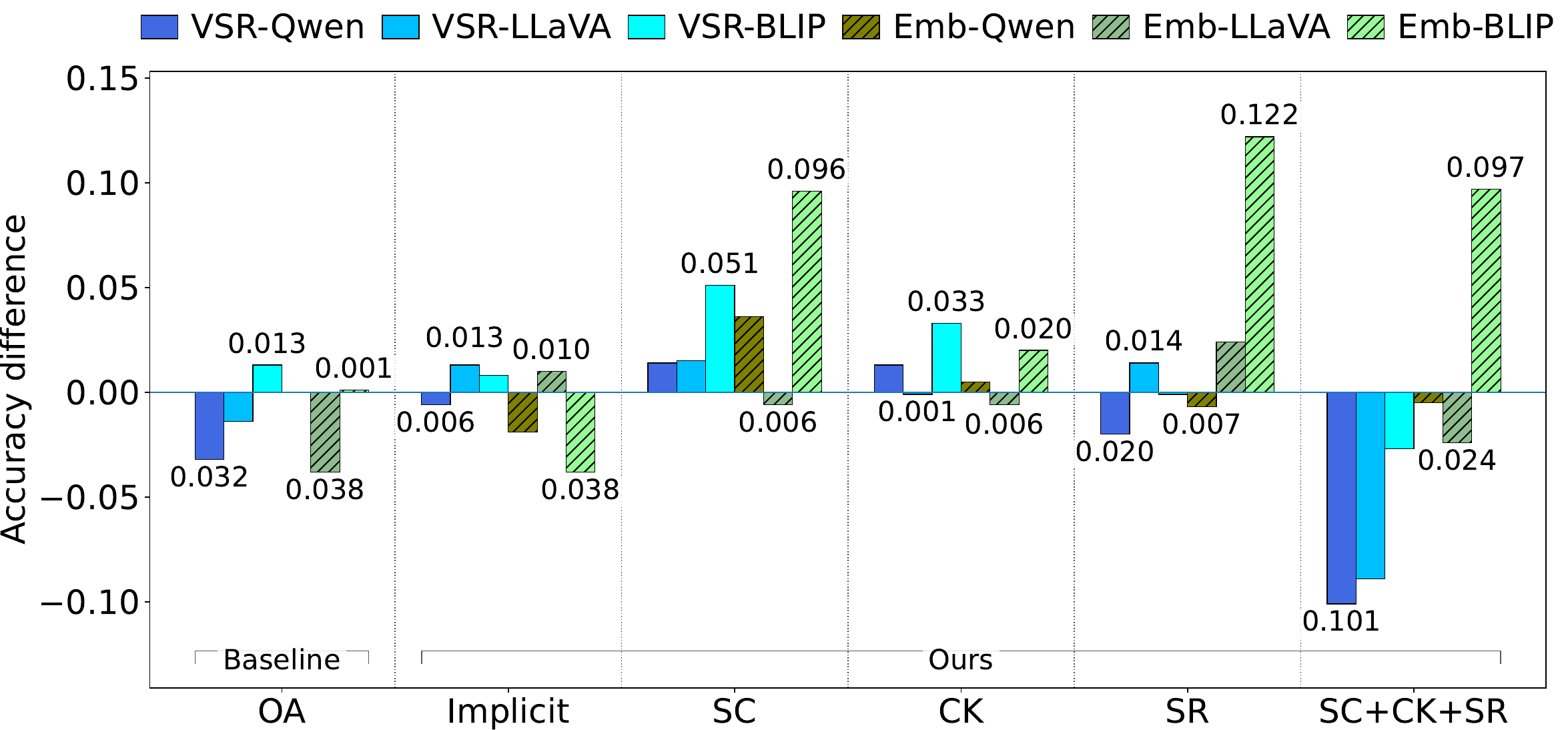}
  \caption{Overall performance comparison between models on VSR and EmbSpatial. The accuracy difference is calculated from the zero-shot performance. OA, SC, CK, and SR stand for Orient Anything, Spatial Contexts, Commonsense Knowledge, and Spatial Reasoning Instructions, respectively.}
  \label{fig:overall_results}
  \vspace{-1.5em}
\end{figure}

We design our experiments to test the hypotheses in Section~\ref{sec:hypo} by analysing how different forms of injected information affect visual spatial reasoning under controlled input interventions.
We treat each VLM as a fixed black-box and vary only the input conditions defined in Section~\ref{sec:intervention}.
We evaluate three representative VLMs: Qwen-2-VL-7B (Qwen) \cite{Qwen2VL}, LLaVA-Next-1.6-7B (LLaVA) \cite{liu2024llavanext}, and BLIP-3-5B (BLIP) \cite{xue2024xgenmmblip3familyopen}, which span diverse architectural and training paradigms.
Experiments are conducted on two public benchmarks.
The VSR dataset~\cite{liu2023visual} evaluates fine-grained spatial relations across 66 categories, while the EmbSpatial dataset~\cite{du-etal-2024-embspatial} focuses on reasoning in cluttered, object-dense scenes.
Together, they test both explicit spatial grounding and abstract relational reasoning.
All results are reported using accuracy.
Injected spatial and relational information is generated using off-the-shelf perception models: Grounding DINO~\cite{liu2023grounding}, Segment Anything 2~\cite{ravi2024sam2}, Depth Anything v2~\cite{yang2024depth}, and Orient Anything~\cite{wang2024orient}.
These components are used solely to construct input contexts; VLM parameters remain fixed throughout.

%% file: sections/05_results.tex
\section{Results}

\subsection{Overall Performance}
To examine how injected information affects VSR, we compare intervention settings, including Orient Anything (OA)~\cite{wang2024orient}. Figure~\ref{fig:overall_results} reports the best accuracy gains over zero-shot prompting on VSR and EmbSpatial in each setting. Zero-shot includes the image, caption, task description, and answer options. Implicit prompting adds light spatial cues without explicit external information. We reproduce OA following the original procedure.

\noindent\textbf{Testing H1 (Spatial Contexts).}
Injecting explicit spatial contexts yields the most consistent improvements across models and datasets. SC improves accuracy by up to 9.6\% and consistently outperforms OA, which often underperforms zero-shot. This supports \textbf{H1}: targeted spatial grounding is more effective than generic or aggregated cues.

\noindent\textbf{Testing H2 (Commonsense Knowledge).}
The impact of CK is model-dependent. While Qwen and BLIP-3 benefit, LLaVA shows slight degradation (–0.6\%) across datasets, indicating that additional relational information can introduce noise. These results support \textbf{H2}, suggesting that commonsense knowledge must be selectively injected to avoid overload.

\noindent\textbf{Testing H3 (Spatial Reasoning Instructions).}
Spatial reasoning instructions (SR) exhibit divergent effects across datasets.
On EmbSpatial, SR substantially improves performance for LLaVA and BLIP-3 (up to +12.2\%), whereas on VSR it degrades accuracy by up to 2.0\%.
This contrast supports \textbf{H3}, suggesting that CoT-style prompting is helpful primarily in settings where clear premises are provided, but can be detrimental when ambiguity persists and over-reasoning amplifies errors.

\noindent\textbf{Combined Interventions and Cognitive Overload.}
Finally, combining SC, CK, and SR does not lead to additive gains.
In several cases, the full combination (SC+CK+SR) significantly reduces accuracy compared to simpler interventions.
This observation provides further evidence of \emph{cognitive overload}: simultaneously injecting multiple signals increases the integration burden on current VLMs and leads to confusion rather than synergy.
Overall, these results reinforce a consistent theme across hypotheses: for current VLMs, injecting a single, task-relevant spatial context is more effective than aggregating multiple forms of external information.

\begin{figure}[t]
    \centering

    \includegraphics[width=0.8\linewidth]{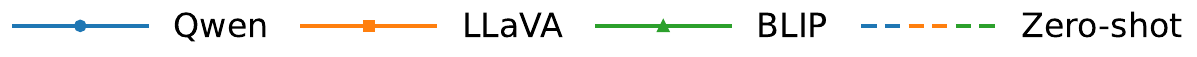}
  
    \begin{subfigure}[t]{0.5\linewidth}
      \centering
      \includegraphics[width=\linewidth]{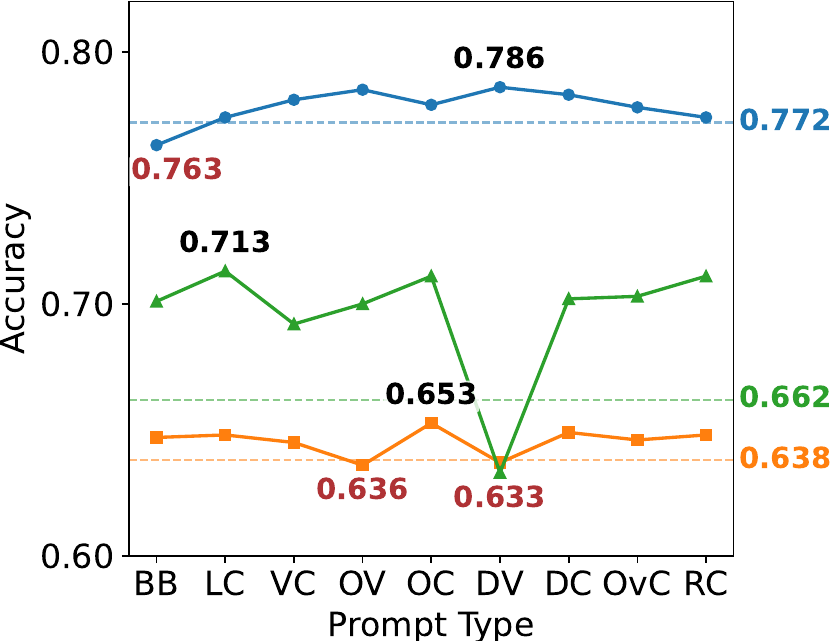}
      \caption{VSR}
      \label{fig:vsr_spatial_context_accuracy}
    \end{subfigure}%
    \begin{subfigure}[t]{0.5\linewidth}
      \centering
      \includegraphics[width=\linewidth]{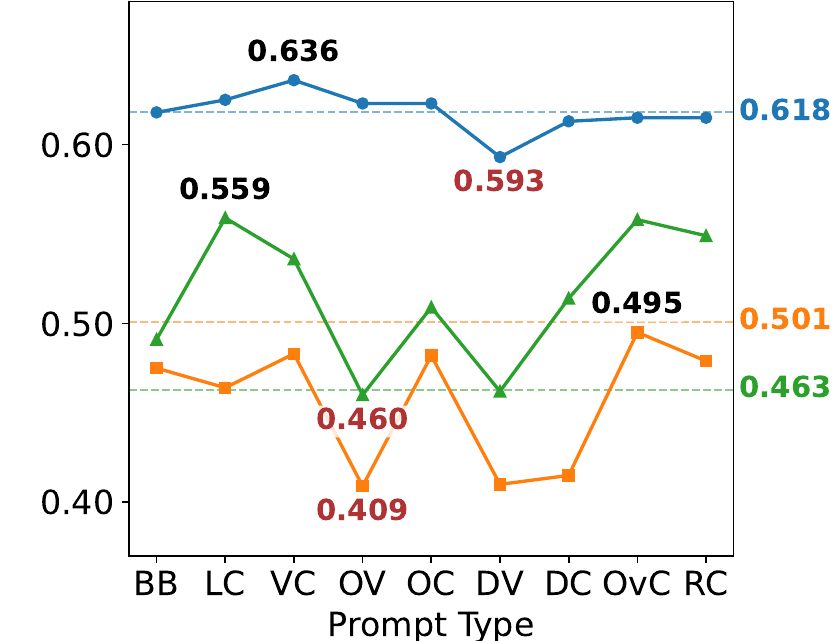}
      \caption{EmbSpatial}
      \label{fig:embspatial_spatial_context_accuracy}
    \end{subfigure}
  \caption{Spatial context overall accuracy (\%). Prompt types are abbreviated as follows (BB: bounding box, LC: lateral context, VC: vertical context, OV: orientation angle, OC: orientation context, DV: metric depth value, DC: depth context, OvC: overlap context, and RC: relative size context).}
  \label{fig:overall_spatial_context_accuracy}
  \vspace{-0.5em}
\end{figure}

\begin{figure}[t]
    \centering

    \includegraphics[width=0.8\linewidth]{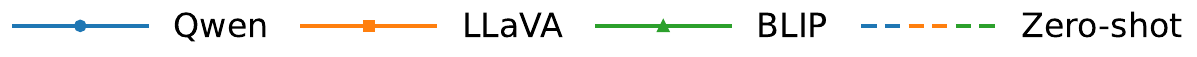}
  
    \begin{subfigure}[t]{0.5\linewidth}
      \centering
      \includegraphics[width=\linewidth]{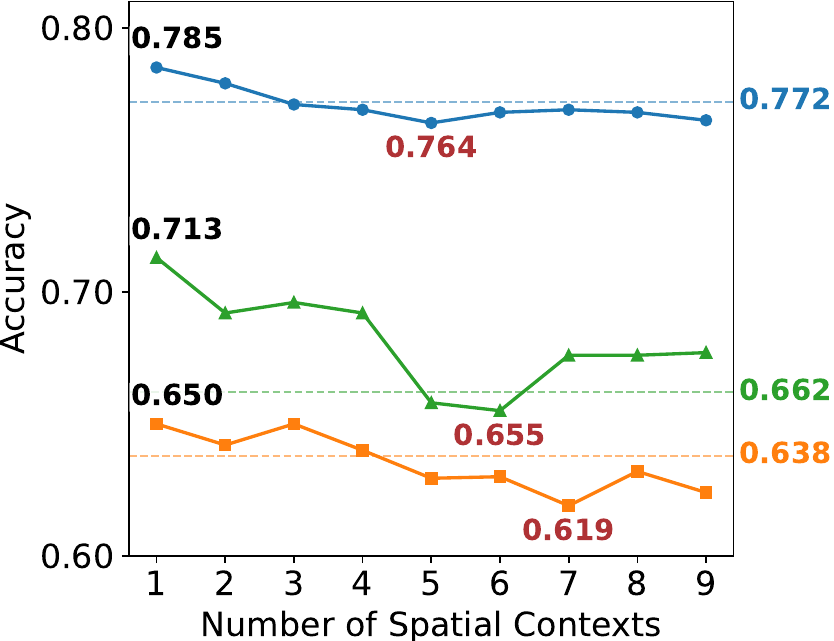}
      \caption{VSR}
      \label{fig:vsr_number_of_spatial_context_accuracy}
    \end{subfigure}%
    \begin{subfigure}[t]{0.5\linewidth}
      \centering
      \includegraphics[width=\linewidth]{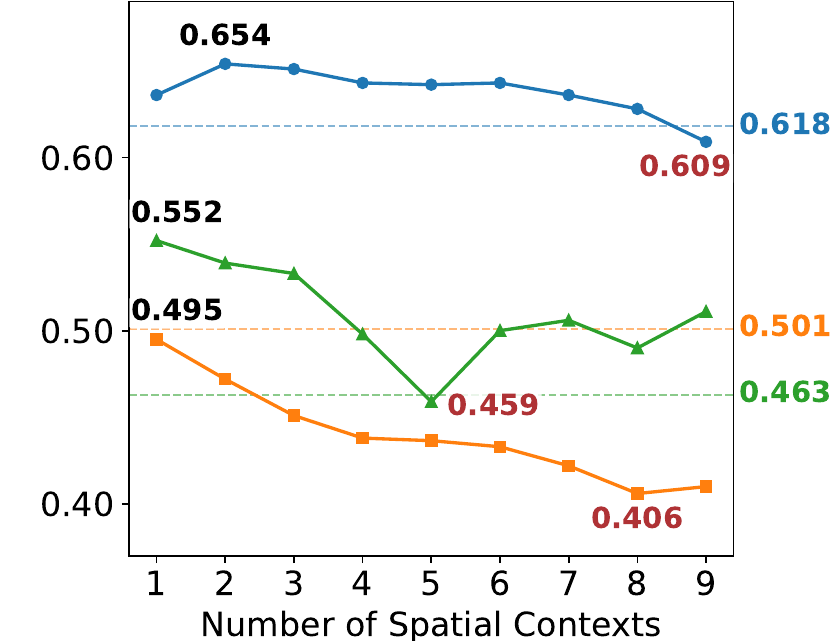}
      \caption{EmbSpatial}
      \label{fig:embspatial_number_of_spatial_context_accuracy}
    \end{subfigure}
  \caption{Performance trend of each model over different numbers of spatial contexts.}
  \vspace{-1em}
  \label{fig:number_of_contexts_vs_accuracy}
\end{figure}

\subsection{Analysis of Spatial Context}
To directly test \textbf{H1}, we examine how individual spatial contexts affect visual spatial reasoning when injected in isolation.
Figure~\ref{fig:overall_spatial_context_accuracy} reports accuracy for each spatial context across models on the VSR and EmbSpatial datasets.
Across both datasets, injecting spatial information generally improves performance over zero-shot prompting; however, the magnitude and direction of improvement vary substantially across spatial context types.
On the VSR dataset, Qwen benefits most from depth-related cues, LLaVA from orientational context, and BLIP-3 from lateral relations.
On EmbSpatial, Qwen performs best with vertical context, LLaVA with overlap relations, while BLIP-3 again favours lateral context.
This variability indicates that spatial contexts are not interchangeable and that their utility depends on both the model and the reasoning demands of the dataset.
The observed differences align with the evaluation focus of each dataset.
VSR probes fine-grained spatial relations across 66 categories, where three-dimensional cues such as orientation and depth are often informative.
In contrast, EmbSpatial evaluates reasoning in cluttered scenes with many irrelevant objects, where two-dimensional relational cues such as lateral and overlap relations are more effective.
These results suggest that spatial context utility is tightly coupled to the underlying spatial abstraction required by the task.
A consistent trend across models and datasets is that verbalised spatial contexts outperform their numerical counterparts.
In particular, raw orientation values and depth values degrade performance for most models (by up to 9.2\%), whereas their verbalised forms lead to stable improvements.
This finding challenges the assumption that more precise, fine-grained numerical inputs are inherently beneficial.
Instead, continuous numeric cues introduce additional integration complexity, requiring models to normalise and interpret values before reasoning.
For current VLMs, this added burden often outweighs the potential informational gain.

\noindent\textbf{Implications for H1.}
Together, these results provide strong evidence for \textbf{H1}.
While spatial context injection can improve VSR performance, not all spatial contexts contribute equally, and increased representational precision does not guarantee better reasoning.
Effective spatial grounding requires selecting a \emph{single, task-relevant} spatial cue and expressing it in a form that the model can reliably consume.
When spatial information is mismatched to the task or overly complex, it is frequently treated as noise.

\begin{figure}[t]
  \centering
  \includegraphics[width=\linewidth]{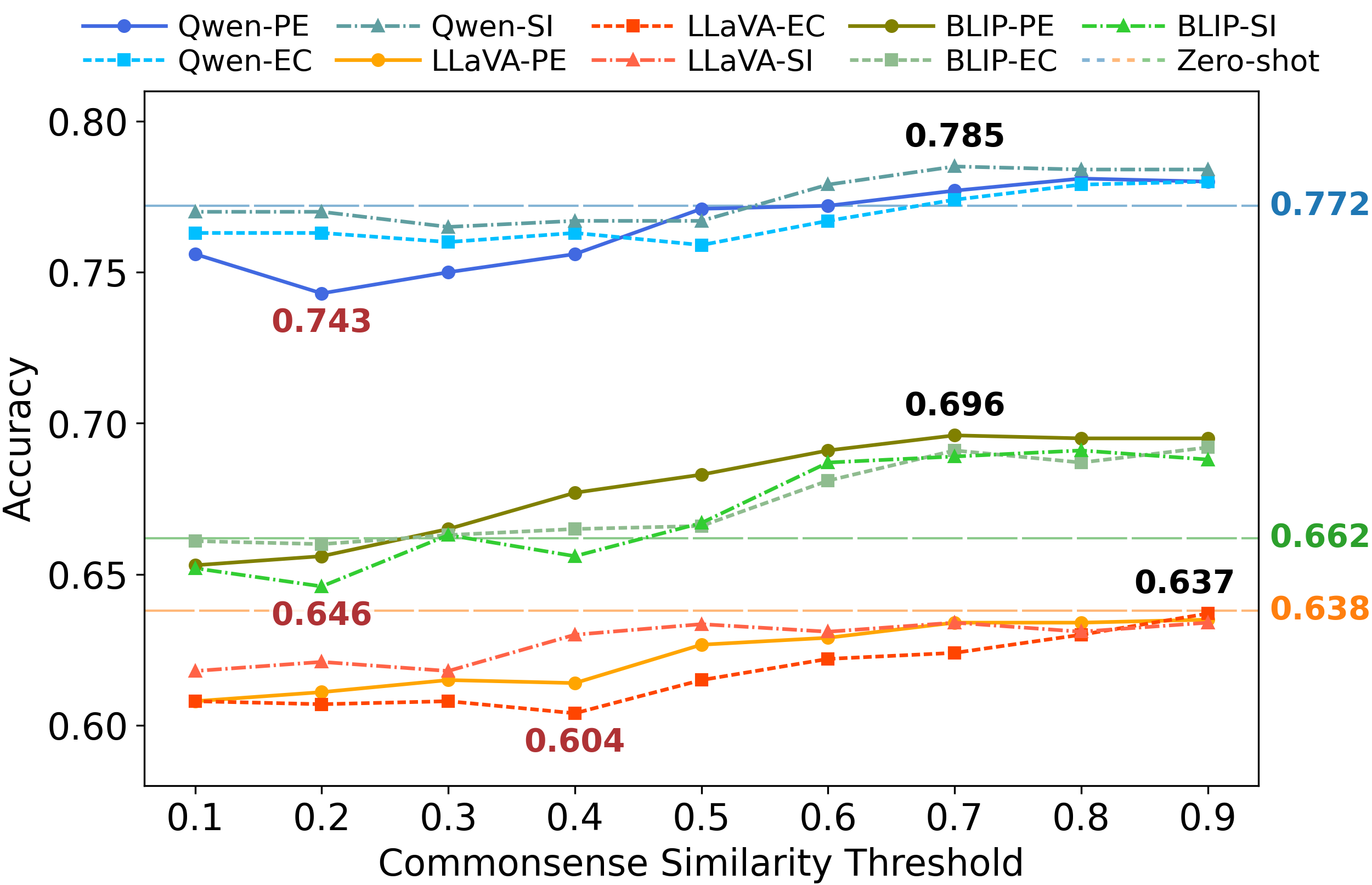}
  \caption{Performance trend of each model over different
thresholds of commonsense knowledge similarity on VSR.}  \label{fig:commonsense_threshold_vs_accuracy_per_category_model}
\vspace{-1.5em}
\end{figure}

\subsection{Effect of Number of Spatial Contexts}
To further test \textbf{H1}, we analyse how performance changes as multiple spatial contexts are aggregated.
Figure~\ref{fig:number_of_contexts_vs_accuracy} shows accuracy trends as the number of injected contexts increases from 0 to 9, added cumulatively in descending order of single-context effectiveness.
Across all three models, performance peaks with a single spatial context and generally declines as more contexts are appended.
A notable drop appears around five contexts, after which accuracy does not recover to earlier levels.
Although minor fluctuations occur, no model achieves higher accuracy with more than four contexts than with three.
These results provide strong support for \textbf{H1}.
Aggregating multiple spatial cues yields diminishing returns and often degrades performance, suggesting that current VLMs struggle to integrate multiple spatial signals simultaneously.

\subsection{Commonsense Knowledge Analysis}
To test \textbf{H2}, we analyse how the amount and type of injected commonsense knowledge affect VSR.
Figure~\ref{fig:commonsense_threshold_vs_accuracy_per_category_model} shows performance as the similarity threshold varies from 0.1 to 0.9 for three knowledge types: Physical-Entity (PE), Event-Centred (EC), and Social-Interaction (SI).
Higher thresholds retain fewer but more relevant statements, whereas lower thresholds admit loosely related knowledge.
Across all models, performance peaks at high thresholds (0.7–0.9) and declines as the threshold decreases.
At suboptimal thresholds (e.g., 0.2 for Qwen and BLIP-3, 0.4 for LLaVA), accuracy falls below zero-shot, indicating that loosely aligned knowledge introduces noise rather than support.
At high thresholds, all knowledge types achieve comparable peak performance, suggesting that diverse but relevant commonsense can assist reasoning.
However, PE knowledge exhibits instability at mid-level thresholds, implying that even spatially aligned commonsense can be harmful when injected imprecisely or excessively.
These findings strongly support \textbf{H2}: commonsense improves VSR only when selectively injected, while excessive or weakly aligned knowledge induces information overload.

\begin{figure}[t]
  \centering

  \begin{subfigure}{\linewidth}
    \centering
    \includegraphics[width=\linewidth]{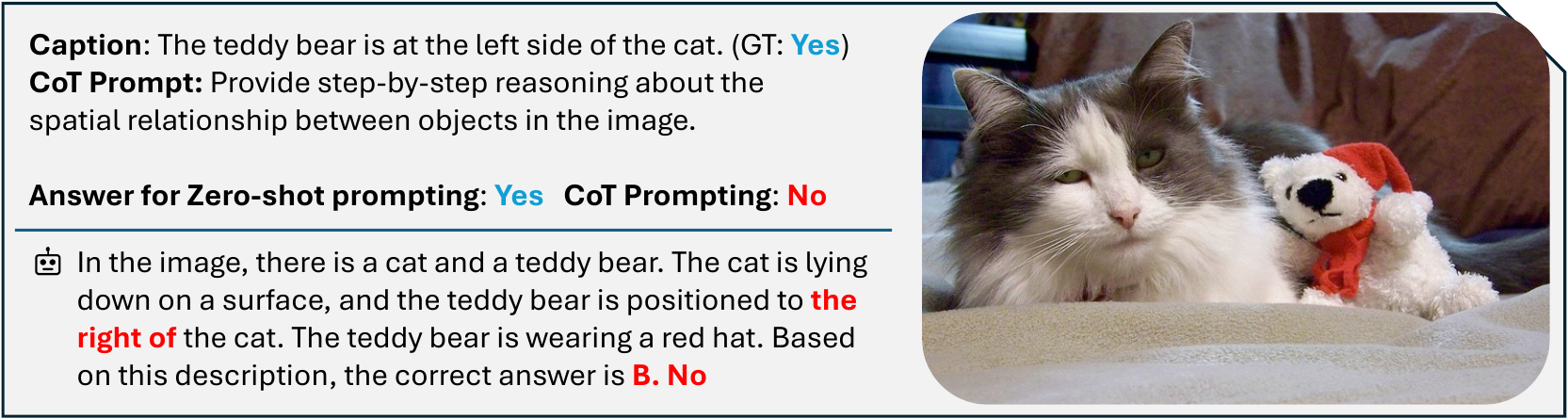}
    \caption{Example of CoT prompting leading to an incorrect answer}
    \label{fig:qualitative_reasoning_intrinsic}
  \end{subfigure}

  \vspace{0.5em}

  \begin{subfigure}{\linewidth}
    \centering
    \includegraphics[width=\linewidth]{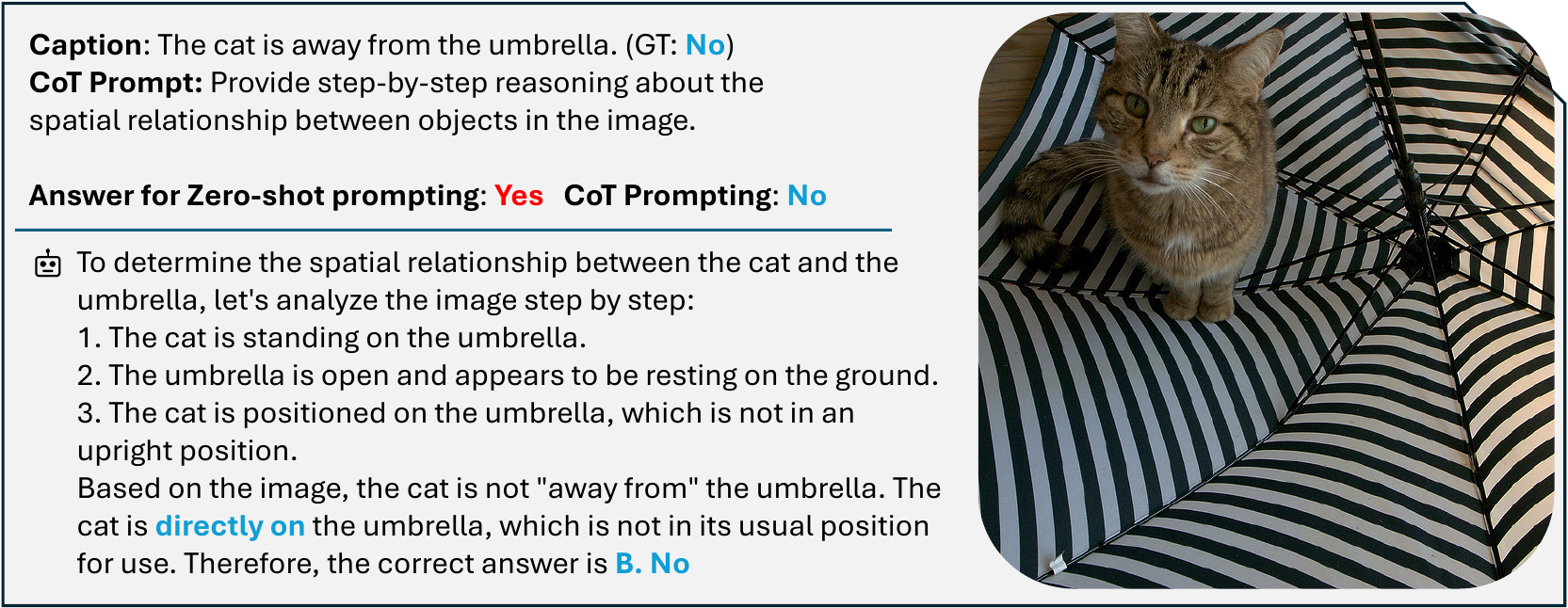}
    \caption{Example of CoT prompting leading to a correct answer.}
    \label{fig:qualitative_reasoning_relative}
  \end{subfigure}

  \caption{Qualitative examples of CoT prompting under different frames.}
  \vspace{-1em}
  \label{fig:qualitative_reasoning}
\end{figure}

\subsection{Impact of Chain-of-Thought Reasoning}
To test \textbf{H3}, we examine when chain-of-thought (CoT) prompting improves VSR and when it amplifies errors.
Figure~\ref{fig:qualitative_reasoning} provides qualitative examples highlighting the role of grounding precision and frame specification.
Figure~\ref{fig:qualitative_reasoning_intrinsic} shows a case with an ambiguous frame of reference.
The teddy bear appears on the right of the cat from the camera’s perspective, but on the left from the cat’s intrinsic perspective.
Because the frame is unspecified, “A is on the right of B” does not entail “A is not on the left of B.”
Under CoT prompting, the model constructs an explicit reasoning chain and incorrectly applies this negation, reinforcing the ambiguity and producing a wrong answer.
By contrast, Figure~\ref{fig:qualitative_reasoning_relative} presents a proxemic distance relation that is frame-independent.
Since the grounding premise is unambiguous (the cat is physically on the umbrella), CoT helps articulate the reasoning and leads to the correct prediction.

\noindent\textbf{Implications for H3.}
These examples provide evidence for \textbf{H3}.
CoT prompting is beneficial only when spatial grounding is sufficiently precise and the underlying inference is unambiguous.
When grounding information is incomplete, frame-dependent, or uncertain, CoT can amplify erroneous assumptions by enforcing a coherent but incorrect reasoning chain.
Therefore, CoT should not be applied indiscriminately in VSR tasks; its effectiveness depends critically on the reliability of the underlying spatial premises.

%% file: sections/06_conclusion.tex
\section{Conclusion}
We conducted a hypothesis-driven analysis of information injection for visual spatial reasoning in vision-language models.
Across three representative VLMs and two benchmarks, our results consistently show that more information does not necessarily lead to better reasoning.
First, spatial context improves performance only when carefully selected; certain representations enhance grounding, while others introduce integration difficulty.
Second, both spatial and commonsense knowledge provide gains only when selectively injected—excessive or weakly aligned information degrades accuracy, revealing clear overload effects.
Third, chain-of-thought prompting is beneficial only under precise and unambiguous spatial grounding; otherwise, it can amplify erroneous assumptions.
Overall, effective VSR therefore requires controlled, task-aligned information injection rather than indiscriminate context accumulation.